# More Practical AI Solution: Breast Ultrasound Diagnosis Using Multi-AI Model Ensemble System

Jian Dai[1], Shuge Lei[1,2], Licong Dong[3], Xiaona Lin[3], Huabin Zhang[4], Desheng Sun[3*], Kehong Yuan[1*]

[1] Graduate School at Shenzhen, Tsinghua University, Shenzhen, China
[2] Computer Science and Engineering, University of South Carolina, SC, United States
[3] Ultrasound department, Shenzhen Hospital of Peking University, Shenzhen, China
[4] Ultrasound department, Beijing Tsinghua Changgeng Hospital, Beijing, China

Corresponding author: Kehong Yuan (yuankh@sz.tsinghua.edu.cn), Desheng Sun (szdssun1@126.com).

**ABSTRACT** *Objective*: Breast cancer screening is of great significance in contemporary women's health prevention. Most current machines embedded in the AI system for breast cancer screening do not meet the clinical criteria. How to make intelligent systems more practical and reliable is a common problem. *Methods*: 1) Ultrasound image super-resolution: the SRGAN super-resolution network reduces ultrasound images' unclearness caused by the device itself and improves the accuracy and generalization of the detection model. 2) We improved the YOLOv4 and the CenterNet models in response to the medical criteria. 3) Multi-AI model: based on the advantages of different AI models, we employ two AI models to detect clinical results via cross-validation; we only accept the same results from both models. *Results*: 1) The YOLOv4 model and the CenterNet model both increased the mAP score by 9.6% and 13.8% respectively, when added the super-resolution model. 2) Two methods for transforming the target model into a classification model are proposed; the unified output is specified to call the multi-AI model. 3) In the classification evaluation experiment, concatenated by the YOLOv4 model (sensitivity 57.73%, specificity 90.08%) and the CenterNet model (sensitivity 62.64%, specificity 92.54%), the multi-AI model will refuse to make judgments on 23.55% of the input data. Correspondingly, the sensitivity has been improved to 95.91% and the specificity 96.02%. *Conclusion*: Our work makes the AI model more practical and reliable in medical image diagnosis, and it performs well in detecting breast lesions. *Significance*: 1) The proposed method makes the target detection model more suitable for remote diagnosis of breast ultrasound image. 2) It provides a new idea for applying AI in medical diagnosis, which can more conveniently introduce target detection models from other fields to serve medical lesion screening.

## I. INTRODUCTION

Breast cancer has become one of the biggest threats to female health, and it is the malignant tumor that caused the highest breast cancer incidence among women[1]. The cumulative risks for females worldwide to age 75 are 5.03%[2]. Early screening and diagnosis are critical factors in reducing the mortality rate of breast cancer[3]. Formally used in clinical practice for decades, breast ultrasonography has gained continuous technological breakthroughs and eventually shows great potential for early screening for its convenience and lower cost, especially in some developing areas such as southern Asian countries[4]. Yet the ultrasonography for early screening is limited for several reasons.

The highly professional training required for ultrasonography operation and diagnosis and the variability of ultrasound images' resolution are among the major handicaps. Ultrasonography relies heavily on the doctor's experience, operating methods, and equipment parameters. The skills of the doctors will significantly influence the diagnosis results[5]. There remains a lack of sonography education and ultrasound-trained physician support in developing countries[6] although access to ultrasound has increased significantly in resource-limited settings. The diagnostic accuracy will be largely lowered by the increasing workloads of sonographers, the fatigue caused by long-term reading, the inexperience of young physicians, etc. Besides, the resolution of ultrasound inspection equipment varied greatly and will also affect doctors' diagnosis. The lower resolution ultrasound equipment generally brings higher misdiagnosis. AI-based computer-aided diagnostic systems can assist the inexperienced physicians and help reduce the workloads, thus showing great promise in assisting breast diagnosis.

There are two major challenges in developing AI-based ultrasound diagnostic systems for breast diagnosis. A significant one is the increasing complexity of algorithms due to the high accuracy requirements for medical diagnosis, which requires the ultra-high computing capability. Yet most current systems use AI-embedded devices and have low computing capability for real-time calculations[7]. Computing in the cloud gained increasing popularity to solve such problem[8].

Another challenge is that the current diagnostic systems always have problems dealing with ultrasound devices with different resolutions. One system usually expertized at a specific range of resolutions of medical images, limiting the systems' generalization and scalability. While the super-resolution model improves the clarity of the image, it also makes the image characteristics more consistent. Besides, most current systems are based on a single traditional detection model, which always performs worse on individual evaluation indicators. To the best of our knowledge, multi-AI models are seldomly employed in breast cancer ultrasound detection.

To address the above challenges in ultrasound breast examination, we develop a more practical ultrasonic computer-aided diagnosis system --- a Multi-AI Combination detection system. We employed the SRGAN super-resolution network to reduce unclear ultrasound images caused by the device and improved the detection model's accuracy and generalization. In addition, the Multi-AI model is cloud-based and has high computing power. Such systems can help unskilled operators improve the accuracy of diagnosis and experienced doctors concentrate on complex images.

II. RELATED WORK

**Super-resolution techniques**. Ultrasound equipment varies in different aspects and among them the resolution size matters most in the lesion detection task. The detection accuracy will decrease when the resolution of input images varies greatly. The goal of Super-Resolution (SR) methods is to recover a High Resolution (HR) image from one or more Low Resolution (LR) input images to the similar resolution. Two kinds of super-resolution methods for medical images are widely used: interpolation-based super-resolution reconstruction algorithm[9] and learning-based super-resolution reconstruction algorithm[10].

SRGAN is the first to reconstruct super-resolution images using a convolutional neural network (CNN). It employed a three-layer CNN to learn the mapping relationship between high-resolution images and low-resolution images to guide the reconstruction[11]. SRGAN presented a super-resolution generative adversarial network, which is the first framework to recover photo-realistic images from $4\times$ down sampling[12]. In the experiment, we use the super-resolution model as part of the image preprocessing to improve the clarity of the image, thereby improving the accuracy of the target detection.

**Object detection model**. The detection model is to detect the breast lesions. The modern detector is usually composed of two parts, a backbone which is pre-trained on ImageNet and a head which is used to predict classes and bounding boxes of objects[13]. According to the difference of the head, it can be divided into two categories: a one-stage object detector and a two-stage object detector. The most representative models for one-stage object detectors are YOLO[12,13,14,15], SSD[16], and RetinaNet[17]. They predict locations and confidences for multiple objects directly based on the whole feature map[18,24]. The two-stage detector combines classifying the box proposals through a CNN model and refining the coordinates with the sliding window manner[19], such as RCNN[20], fast R-CNN[19], faster R-CNN[21]. The recently developed anchor-free one-stage detectors, such as CenterNet[22] and CornerNet[23], detect targets by matching key points rather than the anchor frame principle to improve accuracy and speed. We made some improvements between the head and the backbone on the anchor-free one-stage detectors for the classification.

**Multi-AI model combination.** The Multi-AI model approach does not replace the existing high-precision detection models but combines them together. The existing AI models usually work independently, but reasonable combination of multiple models can improve the classification accuracy. Different models analyze the same picture together, just like an expert consultation, which fully considers the individual evaluation of each model. This is in line with the normal human diagnostic process. We can constantly add more members to make the model more reliable.

What's more, different from general target detection, medical workers do not care too much about indicators such as mAP (defined in Table 2). Doctors only care about whether this tool will misdiagnose patients. To some extent, we hope that AI assistants make a more reasonable judgment, helping doctors identify potential patients instead of displaying a bunch of boxes and scores.

## III. METHOD

### A. DATA AND PREPROCESSING

Common steps of such deep learning systems are as follows: organize the lesion dataset, select the target detection model, adjust the parameters of the model, and transplant it to embedded devices or mobile workstations[24]. The paper also follows such steps.

The dataset in this paper comes from the database of the Ultrasound Imaging Department of Peking University Shenzhen Hospital. We collected 6,860 pictures, including 2,065 images with benign lesions and 3,495 images with malignant lesions, and 1,300 images without lesions.

Table 1 The dataset used in the experiment

| Dataset | Images with benign nodules | Images with malignant nodules | Images without nodules |
|---|---|---|---|
| Training Dataset | 1870 | 3160 | 0 |
| Validation Dataset | 195 | 335 | 1300 |

Five doctors used the open-source software LabelImg to label these data, which were collected from 2804 patients. In order to ensure the accuracy of the data, one doctor marks the image while the other doctor checks it.

The target detection model must have a target during the training process; only the image with a lesion is helpful to the training model. Therefore, as shown in Table 1, we finally selected 5030 lesion images as the training set, and 530 lesion images and 1300 disease-free images as the validation set.

### B. SUPER-RESOLUTION USED IN THE EXPERIMENTS

A clear ultrasound image can improve the clarity of the image and make subsequent analysis easier. When the characteristics of the input images are similar (indexes such as resolution size), the generalization ability of the detection model becomes stronger. In this paper, we use the SRGAN network for super-resolution processing.

The GAN network consists of a generator model (G) and a discriminative model (D). The G model is responsible for generating data that is as close to the real sample as possible, and the D model is used to score the output result, which is to judge whether the generated sample is true or false. As shown in Fig.4, the input of the G model of the SRGAN network is a low-resolution image $I^{LR}$, and the high-resolution output image $I^{SR}$ will be used as the input of the D model together with the original image $I^{HR}$.

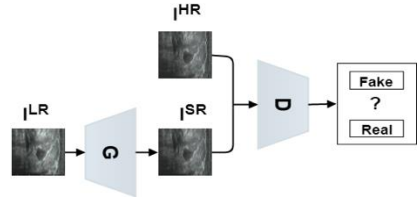

Fig. 4.   The framework of SRGAN

When training SRGAN, it uses a four times downsampling factor to obtain low-resolution images and optimizes in an alternating manner to solve the minimum-maximum adversarial problem:

$$\min_{\theta_C}\max_{\theta_D} \mathbb{E}_{I^{HR}\sim p_{\text{train}}(I^{HR})}[\log D_{\theta_D}(I^{HR})] + \mathbb{E}_{I^{LR}\sim p_C(I^{LR})}\left[\log\left(1 - D_{\theta_D}\left(G_{\theta_G}(I^{LR})\right)\right)\right] \quad (4)$$

Different from previous works, they defined a novel perceptual loss using high-level feature maps of the VGG network[24], combined with a discriminator that encourages solutions perceptually hard to distinguish from the HR reference images[11]. The VGG loss is defined as the Euclidean distance between the feature representation of the reconstructed image $G_{\theta_G}(I^{LR})$ and the reference image $I^{HR}$:

$$l^{SR}_{VGG/i.j} = \frac{1}{W_{i,j}H_{i,j}}\sum_{x=1}^{W_{i,j}}\sum_{y=1}^{H_{i,j}} -(\phi_{i,j}(I^{HR})_{x,y} - \phi_{i,j}\left(G_{\theta_G}(I^{LR})\right)_{x,y})^2 \quad (5)$$

The method of independent alternating iterative training makes the two networks oppose to each other. In the end, the goal of the G network can "deceive" the D network, and the G network is what we need.

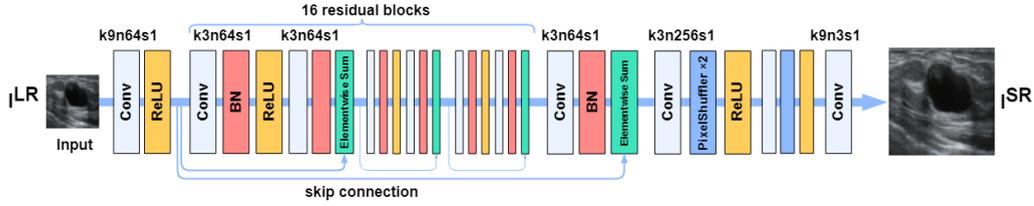

Fig. 5.   The architecture of Generator Network with corresponding kernel size (k), number of feature maps (n), and stride (s) indicated for each convolutional layer

The generative network in the experiment was shown in Fig.5. After the input image passed the convolution block, it went through 16 Residual Blocks (RB) with skip connections. Each RB module consists of two convolutional layers with 3×3 filters and a feature map of 64 channels, followed by a batch-normalization layer and ReLU activation.

The output went through the Elementwise-sum module, and the output of RB was connected to the features of the first convolutional layer. The image can be obtained through 2 upsampling blocks (PixelShuffler) and the last convolutional layer. The scale factor depends on the number (4 used in this article) of upsampling blocks.

*C. THE IMPROVED TARGET DETECTION MODEL*

The super-resolved image will use the detection model to identify the lesion. Considering factors such as the size of the lesion in the entire picture and the uncertainty of the quality of breast ultrasound imaging, the performance of the classification model will be interfered by auxiliary software and imaging effects. This means that we need to clean the data and extract the lesion area before making judgments. Therefore, it is more appropriate for us to use the target detection model.

We need to make some improvements to the detection model. Common approaches are to make some improvements between the head and the backbone[13]. These optimizations will improve the accuracy of the model. As shown in Fig.1, our experiment combines the YOLOv4 model and the CenterNet model to evaluate the pathological images. The principles are similar when combining more models for analysis.

The ultimate goal is to determine whether there is a malignant lesion or not, which is a classification task. The structure of the target detection model and image classification model has many commonalities. Image classification outputs the category information of the picture, while target detection requires a frame to distinguish target objects in each category. When analyzing medical images, these two tasks are performed simultaneously. Because if there is a lesion in a picture, then the patient is ill. In theory, we can use the target detection network to achieve the classification effect.

All in all, there are two areas worthy of optimization: backbone and head. On the one hand, we can connect the output feature map of the backbone part of the YOLOv4 model to the fully connected layer to output category information (as shown in Figure 2). On the other hand, the output of the detection model itself represents information such as the location of the target (lesion), and we can also use the head output information of the CenterNet model for classification (as shown in Figure 3).

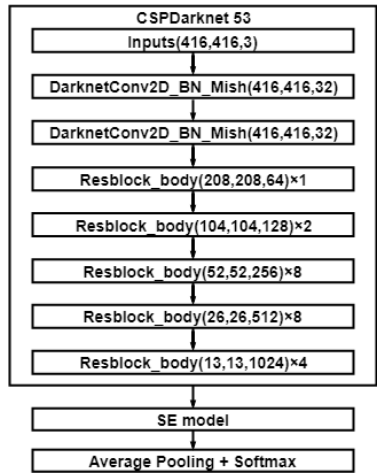

Fig. 2.   The combination of the backbone of YOLOv4 and the SENet

In this paper, we embed the SE module at the output of the Dark net. As shown in Figure 2, the channel of the SE module weights the feature map and inputs it to the fully connected layer to realize the function of the classification network. When the classification network determines that the input picture has a lesion, the remaining network of YOLOv4 determines the location of the target frame.

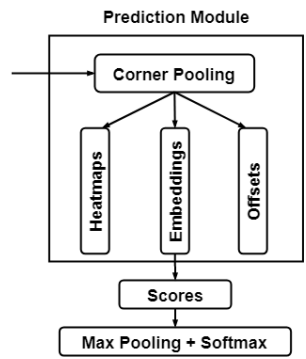

Fig. 3.   The head of CenterNet with classification function

As shown in Figure 3, the CenterNet module will output Heatmaps, Embeddings, Offsets after passing the unique angular pooling[22]. Among them, embeddings contains anchor position and score information in the output of the detection module. We extract the scores in Embedding, which is to score all possible target boxes of each heatmap in the result. Then we use maximum pooling to obtain the maximum probability of disease in different areas of the heatmap, followed by the Softmax to get two categories.

Of course, both improvements of YOLOv4 and CenterNet can be applied to each other. What we have done is to prepare for the classification experiment (as shown in Figure 1).

*D. THE MULTI-AI MODEL ENSEMBLE SYSTEM*

Burnout in radiologists is a significant problem[25]. Intensely concentrating on radiology images for a long time is one of the major causes of burnout. Also, doctors usually have too many workloads due to the large volume of patients and limited doctors. Many AI lesion detection systems require doctors to view real-time ultrasound images of each patient, whether there is a lesion or not. We consider screening out part of the images so that doctors do not need to review those images. This means those being screened out will be judged by the AI system without doctors' supervision. The critical point to achieve such function is the high reliability of the system.

Multi-AI model is unique in that it does not evaluate all pictures, which makes the judgments made by the Multi-AI model more reliable. The super-resolution sub-model and all target detection sub-models are part of the Multi-AI model. How to combine the target detection models is the key to the problem.

As shown in Figure 1, our Multi-AI model does not mark all pictures. Only when the two sub-models agree on their opinions will they give a definite evaluation; the others will be left to the doctors for judgment. Therefore, the output is the affirmative result of multiple models, which largely reduces the misdiagnosis and increase the reliability. We define PP as diseased, NN as disease-free, and PN as unable to judge.

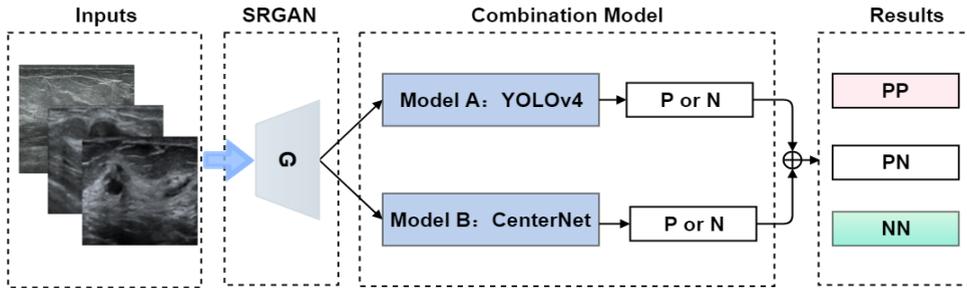

Fig. 1. The framework of the multi-AI model. The input picture pass through the super-resolution model——generator network of SRGAN. Then the images are judged by the "combination mod," and the final output is PP, PN, NN.

Here we averagely weighht the output of the two sub-models as shown in Eqn. (1), and P is the evaluation score of the AI model:

$$P_{multi-AI} = P_{yolov4} + P_{centernet} \quad (1)$$

After adjusting the appropriate threshold ($C_1$ and $C_2$) according to Eqn. (2), the model can achieve the function of Figure 1.

$$\begin{cases} if \ P_{multi-AI} \gg C_1 \ , \ Result = PP \\ if \ P_{multi-AI} \ll C_2 \ , \ Result = NN \\ \quad others \quad \quad \ , \ Result = PN \end{cases} \quad (2)$$

Of course, we can also try the following operations in a joint system composed of multiple models:

$$P_{multi-AI} = \sum_{i=1}^{n} \lambda_i P_i \quad (3)$$

In E qn. (3), $\lambda_i$ is the reliability coefficient of the sub-model. Adjusting the weights of different sub-models can make the model more flexible and stable. Combined with Eqn. (2), it is also very convenient to adjust n to n+1 in a multi-AI model.

We have analyzed how to use a target detection network to complete the classification task, and we will implement similar functions in this section. Our purpose is to carry out preliminary screening of patients, under these circumstances, the multi-model pays more attention to the existence rather than the location of the lesions.

*E. METRICS*

The purpose of the experiment is to improve the clarity of ultrasound images and improve the detection accuracy of the target detection model. Therefore, the evaluation indicators used in the experiments are mainly from the field of target detection and digital imagery.

As shown in Table 2, in order to evaluate the effect of lesion feature detection, we used four indexes: precision, recall, sensitivity, and mAP. As shown in the calculation formula, TP represent the number of positive samples inferred as positive; FN epresent the number of negative samples inferred as positive. Of course, TN and FP have the same logic. Then Precision reflects the model's ability to accurately identify lesion images, and specificity reflects the model's ability to accurately identify disease-free images. If we draw a curve composed of recall score and precision score on the coordinate axis. We can get a area of the curve, which represents the AP value. mAP is average of diferent kinds of AP and usually used to comprehensively measure the accuracy of the detector model.

Table 2 The metrics used in the experiments

| Evaluation index | Method of calculation |
|---|---|
| Precision | $\frac{TP}{TP + FP}$ |
| Recall | $\frac{TP}{TP + FN}$ |
| Specificity | $\frac{TN}{TN + FP}$ |
| AP | Area covered under PR curve |
| MSE | $MSE = \frac{1}{H \times W} \sum_{i=1}^{H} \sum_{j=1}^{W} (X(i,j) - Y(i,j))^2$ |
| PSNR | $PSNR = 10 \cdot \log_{10}(\frac{MAX_I^2}{MSE})$ |
| SSIM | $SSIM(x,y) = \frac{(2\mu_x\mu_y + c_1)(2\sigma_{xy} + c_2)}{(\mu_x^2 + \mu_y^2 + c_1)(\sigma_x^2 + \sigma_y^2 + c_2)}$ |
| RTP | The task reduced by AI machine |

The classification task is not the same as the target detection task. It only makes a judgment on whether there is a malignant lesion or not, rather than predicts the location of the lesion in the ultrasound image. Therefore, our experiment is divided into two tasks: improve the accuracy of the detection model first, and then use a better network to implement the classification task. We have discussed the specific implementation in section III.B.

We also defined an RTP indicator (Reduced Task Proportion), which means the reduced workload that the AI-assisted diagnostic device contributed. The Multi-AI Model does not make an evaluation for every input picture. Only when the evaluations of all sub-models are consistent, the system will give its judgment, and the other pictures will be given to the doctor. Since it only selects certain pictures to make judgments, we use RTP indicators to comprehensively evaluate the practical significance of the model. Of course, the higher these indicators, the more practical our work is.

MOS indicators were introduced as subjective evaluation indicators.

Mean Opinion Score (MOS): the observer makes a subjective evaluation of the image's quality. The larger the value (between 0 and 5), the better the image quality. We compared the evaluation value of the algorithm with the subjective score value from human judges.

## IV. RESULTS

*A. SUPER RESOLUTION*

A clearer image plays an important role in subsequent processing. We need to verify that super-resolution can improve the quality of ultrasound images and improve the effect of lesion detection. In order to verify the actual effect of the super-resolution model, we evaluate both the image quality itself and the improvement of the performance of the target detection model.

The training and testing of super-resolution models require low-resolution ultrasound images. The low-resolution image is obtained by down-sampling high resolution image; the reasons are as follows: first, most general SR datasets are constructed based on down-sampling operations with a scale factor of four. The commonly used down-sampling operation is the cubic bilinear interpolation with anti-aliasing[26]; second, it is difficult for different ultrasound equipment to scan the same section at the same angle in the actual scenario. Also, it is meaningless for doctors and patients in the ultrasound department to use cheap equipment to match the parts that have already acquired high-definition ultrasound images. If the angle or position of the low-resolution section in the dataset is not appropriate, it will have a great impact on the experiment.

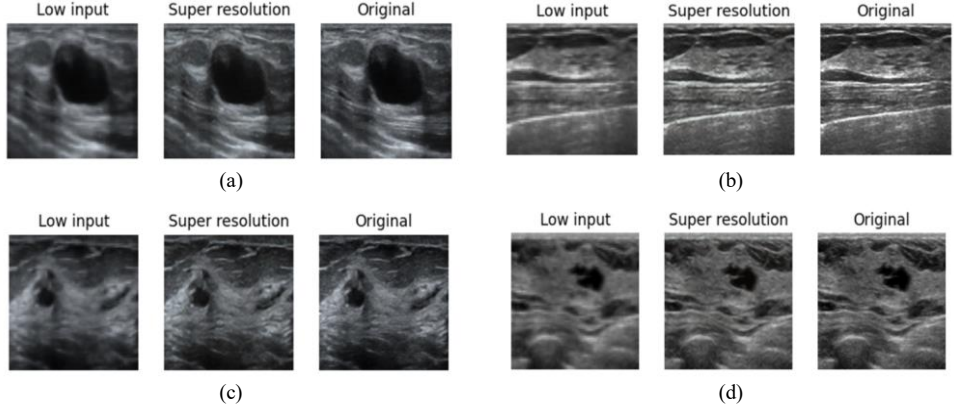

Figure 6 Super-resolution breast ultrasound images

There are three main types of the data: the lesion image (a), the normal image (b), and the suspected lesion image (c) (d). In the super-resolution model, the size of the image will change, and we need to observe them separately. We finally chose three different image sizes: the sizes of "Low input," "Super-resolution," and "Original" are 64*64, 256*256, 720*1024, respectively.

Table 3 Evaluation score of image super-resolution

| Size | 720×1024 | 256×256 | 64×64 |
| --- | --- | --- | --- |
| PSNR | 23.53 | 24.48 | 26.97 |
| SSIM | 0.5523 | 0.5592 | 0.8703 |
| MOS | 3.67 | 3.98 | 4.12 |

We also analyzed the score of the image in three different sizes, and the output image will be the input data of the next target detection model. When evaluating images through qualitative indicators such as PSNR and SSIM, it is necessary to ensure that the contrast images are consistent in size.

As shown in Table 3, if we reduce output image size, the evaluation index score will become higher. However, the smaller the size of the image, the more difficult it is to detect the location of the lesion. In this paper, we use the output with a size of 256×256 as the input of the next target detection model.

### B. TARGET RECOGNITION

It is necessary to use the target detection network to verify the accuracy of the super-resolution image. That is to analyze the model's score changes from the three types of image resolution sizes.

In order to avoid the contingency of a single model, we use both YOLOv4 and CenterNet to verify the actual model effect. As an anchor-based model, YOLOv4 added an artificial prior distribution; thus, the predicted value variability during training is relatively small, which makes anchor-based networks easier to train and more stable. As a key-point-based method, CenterNet has a larger and more flexible solution space. Getting rid of the amount of calculation brought by the use of anchors makes detection and segmentation further towards real-time high precision.

As shown in Table 4, various indicators (precision, recall, and mAP) all indicate that although the super-resolution generated image has a certain distance from the original image, it is significantly better than the low-resolution image. The accuracy improvement effect on CenterNet is better than YOLOv4, which also shows that the key-point-based model is more dependent on the high resolution of the image than the anchor-based network.

Table 4 The evaluation index used in our experiment

| Model | Evaluation index | LR | SR | HR |
|---|---|---|---|---|
| YOLOv4 | Precision | 58.5% | 60.5% | 65.5% |
|  | Recall | 65.0% | 73.6% | 79.5% |
|  | mAP | 60.7% | 70.3% | 76.4% |
|  | Specificity | 84.5% | 89.5% | 90.1% |
| CenterNet | Precision | 51.6% | 63.0% | 68.4% |
|  | Recall | 52.9% | 59.2% | 62.5% |
|  | mAP | 58.1% | 71.9% | 76.3% |
|  | Specificity | 83.8% | 82.6% | 92.5% |

Moreover, the Specificity value of the model is significantly higher than the Precision value as we put more lesion-free images in the test data. We need to avoid increasing the false alarm rate while reducing missed inspections. That is to reduce the overall error rate in Multi-AI model ensemble system.

The ultimate goal is to make the Multi-AI model embedded system perform better, not to optimize the target detection sub-model. The experiments showed that the super-resolution can significantly improve the performance of existing target detection models. It is worth noting that the super-resolution model works during the detection rather than in training the model. This means it won't influence the subsequent multi-model combination strategy.

*C. MULTI-AI MODEL ENSEMBLE SYSTEM*

The system pays more attention to the reliability and practicality of the model. We have established a remote ultrasound diagnosis server to transmit the collected data to the cloud in real time for analysis. This makes the multi-AI model pay less attention to computing speed and resources.

We only pay attention to the accuracy of the pictures that the model can make judgments, because doctors will analyze the images that cannot be judged. Thus, we discarded some important indicators in the target detection, such as sensitivity, recall rate, etc.

Our results are expressed by the reproduction matrix, which is composed of TP, FP, TN, and FN. The result of a single classification model conforms to the following rules: the sum of TP and FN is 530, and the sum of FP and TN is 1300. Considering the relationship between multi-AI model and single improved detection model, their results have the following limitations:

$$\begin{cases} TP_{multi-AI} \leq TP_{yolov4} \cap TP_{centernet} \\ TN_{multi-AI} \leq TN_{yolov4} \cap TN_{centernet} \\ FP_{multi-AI} \leq FP_{yolov4} \cap FP_{centernet} \\ FN_{multi-AI} \leq FN_{yolov4} \cap FN_{centernet} \end{cases} \quad (6)$$

The Multi-AI model makes trade-offs from the intersection of the sub-models' outputs. For this purpose, we set two thresholds in Eqn. (2), which make the model selection more sensible.

In order to test the actual effect of multi-AI model individually, we use original image of the ultrasound dataset as the input for this experiment. The results of the confusion matrix is as follows:

Table 5 The confusion matrix of the models

A) YOLOv4

| Prediction \ Label | Positive | Negative |
|---|---|---|
| Positive | 306 | 129 |
| Negative | 224 | 1171 |

B) CenterNet

| Prediction \ Label | Positive | Negative |
|---|---|---|
| Positive | 332 | 97 |
| Negative | 198 | 1203 |

C) YOLOv4+CenterNet

| Prediction \ Label | Positive | Negative |
|---|---|---|
| Positive | 305 | 43 |
| Negative | 13 | 1038 |

The images with benign and malignant tumors were treated as negative samples and those without lesions as positive samples. As shown in the two-class confusion matrix in Table 5, the performance are relatively poor when using the feature map of the backbone of a single target detection model for classification.

The results of single model show that there is an overfitting phenomenon in detection. Single classification model is prone to the missed detection, which makes the FN value too large. What's more, the model trained with target detection indicators will take more into account the images with multiple lesions. Therefore, evaluating the results with classification indicators shows poor performance.

In the experiment, adjusting the threshold will influence the multi-AI model's performance. Specifically, the stricter the threshold score, the less evaluation the multi-AI model can make; then the error rate can be lower than single model, and vice versa. In actual situations, people who are misjudged as having the disease will receive doctors' second examination, while those who are misjudged as disease-free will not, in which the latter case will badly influence patients' health and should be avoid as much as possible. Therefore, in the multi-AI model experiment, the number of positive images judged to be disease-free should be as few as possible.

In order to make the experimental results clearer, we define the sensitivity and specificity as shown in Eqn. (7):

$$\begin{cases} Sensitivity_{multi-AI} = \frac{TP_{multi-AI}}{TP_{multi-AI}+FP_{multi-AI}} \\ Specificity_{multi-AI} = \frac{TN_{multi-AI}}{TN_{multi-AI}+FP_{multi-AI}} \end{cases} \quad (7)$$

The results were presented as follows:

Table 6 The metrics used in the classification experiment

| Model | Sensitivity | Specificity | RTP |
|---|---|---|---|
| YOLOv4 | 57.73% | 90.08% | 100% |
| CenterNet | 62.64% | 92.54% | 100% |
| YOLOv4+CenterNet | 95.91% | 96.02% | 76.45% |

From Table 5 and Table 6, we can see that the model is more conservative in the positive judgment of the disease, which makes sensitivity low and specificity high relatively. This will reduce the correct positive diagnosis, yet it will also reduce the misdiagnosis rate, making the entire model more reliable.

All in all, as shown in the experimental results of a single target detection model, the idea of the joint model allows them to play an effect far exceeding that of a single model, which validates a new idea for medical image detection. The good performance of related indicators shows that the multi-AI model is much more reliable compared to the single target detection model. At the same time, using the detection model with adjusted parameters is more in line with the actual use scenario.

**V. Conclusions and future work**

The super-resolution experiments performed well in imaging quality and detection effects. Yet, the low-resolution images were manually generated rather than collected from the ultrasound equipment, making the model generalization slightly insufficient. The super-resolution model also employed the VGG network. Therefore, this part can also act as the classification or the evaluation tasks as a subsystem.

Although the multi-AI model can reduce the doctors' workloads, as shown in the experiments, the workload would still be very heavy when the input data size is large. In this case, the score of the indicator RPT will decrease. It was assumed that the more "members" in the combination model, the more difficult it is to reach a consistent opinion; thus, we only included two "members". We may make the final decision through a voting system or by multiple members according to their weights[27]. From this perspective, the multi-AI model ensemble system provides a feasible solution and verifies the feasibility.

Algorithm engineers usually pay their efforts to optimize the model, yet less attention to help medical workers reduce their workload. In other words, many systems do not consider doctors' needs. One of this work's contributions is coming up with a practical system with considering the doctors' needs in the actual scenario.

Finally, how the neural network model serves medical images is based on the purpose. Here we improved the detection model to complete the classification task, which is actually a very practical tip. Although many model improvements are based on the combination of channels or feature maps, apart from the partial increase or decrease in the evaluation score, they have not brought much substantial help to doctors. Unless a single model can approach 100% accuracy, it is difficult to bring AI models into practice in medical imaging. Here is how the multi ensemble AI system helps.

**ACKNOWLEDGMENT**

The authors thank Shenzhen Science and Technology Planning Project (NO. JSGG20191129103020960) and Guangdong Cloud Intelligent Ultrasonic Engineering Technology Research Center (No. 2019B092) for supporting this research.